\documentclass[10pt,twocolumn,letterpaper]{article}

\usepackage[pagenumbers]{cvpr}

\usepackage{makecell, graphicx, romannum, amssymb, ccaption, setspace, algorithm, algpseudocode, enumitem, pifont, etoolbox, fontawesome5, longtable, array, arydshln, bbding, tabularx, url, multirow, caption, capt-of}
\usepackage[most]{tcolorbox}

\usepackage{amsmath,amsfonts,bm}

\def\eqref#1{equation~\ref{#1}}
\def\1{\bm{1}}

\DeclareMathAlphabet{\mathsfit}{\encodingdefault}{\sfdefault}{m}{sl}
\SetMathAlphabet{\mathsfit}{bold}{\encodingdefault}{\sfdefault}{bx}{n}

\definecolor{cvprblue}{rgb}{0.21,0.49,0.74}
\usepackage[pagebackref,breaklinks,colorlinks,allcolors=cvprblue]{hyperref}

\renewcommand{\paragraph}[1]{\vspace{1.25mm}\noindent\textbf{#1}}

\setcitestyle{square, comma}

\title{SALOVA: Segment-Augmented Long Video Assistant \\for Targeted Retrieval and Routing in Long-Form Video Analysis}

\author{%
  Junho Kim\thanks{Equal contribution. $\dagger$ Corresponding author.}~~~~~~~~~~Hyunjun Kim\footnote[1]{}~~~~~~~~~~Hosu Lee~~~~~~~~~~Yong Man Ro\footnote[2]{}\\
Integrated Vision and Language Lab, KAIST, South Korea \\
  \tt\small{\{arkimjh,~kimhj709,~leehosu01,~ymro\}@kaist.ac.kr} \\
{\href{https://ivy-lvlm.github.io/SALOVA/}{https://ivy-lvlm.github.io/SALOVA}}
}

\begin{document}
\pagenumbering{arabic}
\maketitle
\begin{abstract}
Despite advances in Large Multi-modal Models, applying them to long and untrimmed video content remains challenging due to limitations in context length and substantial memory overhead. These constraints often lead to significant information loss and reduced relevance in the model responses. With the exponential growth of video data across web platforms, understanding long-form video is crucial for advancing generalized intelligence. In this paper, we introduce \textbf{SALOVA}: \textbf{S}egment-\textbf{A}ugmented \textbf{LO}ng \textbf{V}ideo \textbf{A}ssistant, a novel video-LLM framework designed to enhance the comprehension of lengthy video content through targeted retrieval process. We address two main challenges to achieve it: (\lowercase\expandafter{\romannumeral1}) We present the SceneWalk dataset, a high-quality collection of 87.8K long videos, each densely captioned at the segment level to enable models to capture scene continuity and maintain rich descriptive context. (\lowercase\expandafter{\romannumeral2}) We develop robust architectural designs integrating dynamic routing mechanism and spatio-temporal projector to efficiently retrieve and process relevant video segments based on user queries. Our framework mitigates the limitations of current video-LMMs by allowing for precise identification and retrieval of relevant video segments in response to queries, thereby improving the contextual relevance of the generated responses. Through extensive experiments, SALOVA demonstrates enhanced capability in processing complex long-form videos, showing significant capability to maintain contextual integrity across extended sequences.
\end{abstract}    
\section{Introduction}
\label{sec:intro}

Recent advancements in Large Language Models (LLMs)~\cite{chatgpt, gpt4, Gemini} have brought us one step closer to achieving Artificial General Intelligence (AGI). Next following step, the current trend is shifting toward modular systems that integrate various multi-modality, leveraging the exceptional generalization and reasoning capabilities of LLMs to evolve into Large Multi-modal Models (LMMs). Accordingly, users can unrestrictedly interact with the models across various modalities beyond text, expanding the scope of machine understanding and enhancing user engagement. Especially, considering the widespread adoption of long-form videos across various web platforms, the importance of understanding long, untrimmed video has become increasingly prominent in the multi-modal domain.

After the pioneer works~\cite{liu2023visual, liu2023improved, dai2023instructblip} utilizing visual instruction tuning to augment vision perception into LLMs, remarkable strides~\cite{dong2024internlm, ye2023mplug, chen2023internvl} have been made in aligning cross-modal consistency\textemdash especially between vision and language domains. Albeit more recent models~\cite{zhang2024internlm, li2024llava} integrate various vision modalities all at once, current approaches still face significant challenges in understanding untrimmed and long-form video content. The main challenge is attributed to the limited context length of LMMs, which is an inherent structural limitation that restricts the models to process only a finite number of tokens as the input sequences. We can exemplify that LLaVA series~\cite{li2024llavanext, li2024llava}, when processing a video data, require $144$ visual tokens per each frame, where numerical approximation is only maximum of $\sim$$56$ frames using $8$K max context length LMMs, which is still limited to handle long sequence data. 

Accordingly, current video-LMMs~\cite{lin2023video, cheng2024videollama, liu2025st} rely on (\lowercase\expandafter{\romannumeral1}) sparse frame sampling to represent entire videos~\cite{zhang2024beyond, jin2024chat}, (\lowercase\expandafter{\romannumeral2}) dense compression of visual tokens into a smaller size to manage the excessive number of frames~\cite{maaz2023video, li2025llama}, and (\lowercase\expandafter{\romannumeral3}) adaptive pooling strategies~\cite{xu2024pllava, xu2024slowfast} based on the SlowFast approach~\cite{feichtenhofer2019slowfast}, all aimed at fitting the long video sequences within the limited context window of LMMs. Several studies focusing on the long video understanding task have presented memory-augmented generation~\cite{he2024ma, song2024moviechat} utilizing an additional buffer to embed long-term information, or have extended the context using RoPE-based frequency extension during the training~\cite{zhang2024long}. Despite of such endeavors, when handling massive video frames, previous works still confront restricted context size and significant memory overhead, which leads to substantial visual information loss. As critical events may be overlooked by the models, this hinders their ability to fully capture context changes in lengthy videos, resulting in inaccurate and irrelevant responses for the user queries.

Starting from the intuitive insight outlined below, in this paper, we propose a retrieval-driven approach for long video understanding with LLMs. Analogous to the recent Retriever-Augmented Generation (RAG) systems~\cite{lewis2020retrieval} (widely adopted in LLMs), which retrieve relevant information from external factual knowledge, humans naturally employ similar strategies when seeking specific information, efficiently locating and referring necessary materials to answer targeted questions\textemdash \textit{e.g.,} imagine that we are taking open-book exams or searching for a certain recipe in a cookbook. Given a long and untrimmed video, mirroring the targeted retrieval processes, we introduce a novel framework, \textbf{S}egment-\textbf{A}ugmented \textbf{LO}ng \textbf{V}ideo \textbf{A}ssistant (\textbf{SALOVA}) to effectively handle the long sequence visual inputs by retrieving the relevant video segments. 

To construct our video segments retrieval framework, central challenge hinges on establishing two main components: (\lowercase\expandafter{\romannumeral1}) Densely captioned video data, which consists of video-caption pairs with progressively-captioned descriptions that change throughout each video used to train the model to accurately identify relevant video segments. (\lowercase\expandafter{\romannumeral2}) Dynamic routing mechanism, which selects pertinent video segments for the queries, followed by being connected to LLMs. To address that, our approach is outlined as follows.

\begin{figure*}[t!]
\centering
\includegraphics[width=1.0\textwidth]{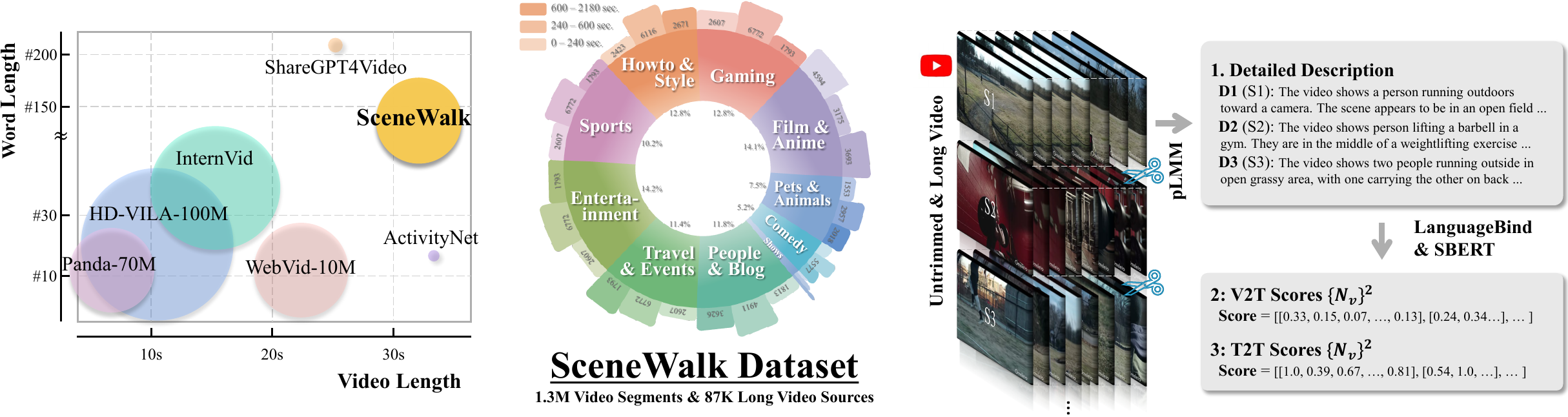}
\vspace{-0.8cm}
\begin{flushleft}
    \hspace{0.3cm} (a) Video-Text Dataset Comparison \hspace{0.5cm} (b) SceneWalk Statistics \hspace{1.4cm} (c) Overall Pipeline for Data Collection   
\end{flushleft}
\vspace*{-0.5cm}
\caption{The overview of the SceneWalk dataset includes (a) dataset comparison, (b) detailed statistics, and (c) the annotation pipeline for description and score collection. Note that the scale of circles in \cref{fig:1}(a) indicates the data size, and the color distribution in \cref{fig:1}(b) denotes the video duration in each video category\textemdash brighter colors correspond to shorter video durations. Further details about the dataset are provided in Appendix A.}
\vspace*{-0.3cm}
\label{fig:1}
\end{figure*}

\paragraph{Data. (\S\ref{sec:data})} Recently several video-text paired datasets~\cite{caba2015activitynet, bain2021frozen, xue2022advancing, wang2023internvid, chen2024panda, chen2024sharegpt4video} have been released, but they are inadequate for handling long and untrimmed video data, where only partial video moments are described with limited word length as compared in \cref{fig:1}(a). To handle such insufficiency of detailed descriptions within the videos and the short durations of both videos and texts, we introduce the \textbf{SceneWalk} dataset, a new high-quality video dataset with thorough captioning for each video. It includes dense and detailed descriptions for every video segment across the entire scene context. The SceneWalk dataset, sourced from long and untrimmed $87.8$K YouTube videos (avg. $486$ seconds each), features frequent scene transitions across a total of $11.8$K hrs video duration and $1.3$M massively segmented video clips. Each video segment in the dataset is provided with a detailed description (avg. $137.5$ word length), generated by combining pre-trained models~\cite{zhulanguagebind, thakur-2020-AugSBERT} and manual curation from human.

\paragraph{Architecture. (\S\ref{sec:method})} Utilizing the constructed video dataset, SALOVA learns to identify relevant video segments for the given queries within each video source and then auto-regressively predicts the next token. To do so, we present two architectural designs to seamlessly incorporate the retrieved segments in an end-to-end training: Spatio-Temporal Connector and Segment Retrieval Router. By focusing on the relevant segments, our framework can perform deeper reasoning without being constrained by context length limitations. Additionally, we present FocusFast approach, which intensively analyzes the selected segments for detailed comprehension (focus pathway), while quickly accessing overall contextual information with routing tokens obtained from the entire video segments (fast pathway). The strategy ensures SALOVA to maintain comprehensive video understanding while prioritizing details where it is most needed, effectively enhancing long and untrimmed video interpretation.

Through extensive experiments and analyses, we corroborate that competitive performance of SALOVA to the existing video-LMM models in understanding complex long-form videos. Also, our results show significant reductions in the loss of crucial visual information and a lower risk of omitting important events, demonstrating the effectiveness of our proposed method across various video benchmarks.

Our contribution can be summarized into three-fold:
\begin{itemize}
\item We introduce the \textbf{SceneWalk} dataset, a high-quality and densely-captioned video dataset with detailed segment-level descriptions from $87.8$K long and untrimmed video sources. The proposed dataset provides rich context and scene continuity, enabling effective training for long-form video understanding.

\item We propose Segment-Augmented LOng Video Assistant (\textbf{SALOVA}), a novel video-LMM framework designed to enhance long video comprehension by targeting relevant video segments in lengthy videos, optimizing the model’s focus on essential segment targets for the given queries.

\item Through extensive evaluation, we validate that SALOVA improves overall long video understanding capabilities by effectively integrating relevant video segments, thus optimizing to handle long and untrimmed video content.
\end{itemize}
% %################################################################################
% Figure
% %################################################################################

\section{Related Work}
\label{sec:formatting}

\subsection{Large Multi-modal Models}
After the emergence of LLMs~\cite{brown2020language, touvron2023llama}, which can actively interact with users through back-and-forth conversations, as a next leap, various research efforts~\cite{alayrac2022flamingo, li2023blip, huang2024language} integrate different modalities into the LLMs, utilizing their core reasoning and zero-shot capabilities. Building on the open-sourced models~\cite{touvron2023llama, vicuna}, seminal works~\cite{liu2023visual, ye2023mplug, dai2023instructblip} have bridged image and text modality under the visual instruction tuning and presented multi-modal assistant models that possess visual perception and QA capabilities. Since then, numerous research studies have been introduced to (\lowercase\expandafter{\romannumeral1}) enhance vision understanding with advanced architectures~\cite{dong2024internlm} or higher resolutions~\cite{liu2023improved, li2024llavanext}, (\lowercase\expandafter{\romannumeral2}) implement more sophisticated alignment layers~\cite{cha2023honeybee, mckinzie2024mm1} between modalities, and (\lowercase\expandafter{\romannumeral3}) train the models with more high-quality data and larger model parameters.

Recent focus has shifted towards more unified modality processing following the release of omnivorous models~\cite{gpt4o}. Some recent omni-versions of LMMs~\cite{ye2024mplug, li2024llava} can handle combinatorial subsets from various modality sources, such as images, videos, audio, speech, and depth. However, the current LMMs for video~\cite{lin2023video, maaz2023video, li2024mvbench} still lack of capturing the necessary details to effectively process video information due to their sparse frame sampling strategy. While such approach is seemingly adequate for relatively shorter videos, it may fail to capture comprehensive spatio-temporal information, potentially compromising the accuracy of model responses to user queries. In this paper, SALOVA first retrieves relevant video segments, then concentrates on more granular video cues. Such targeted focus allows the model to effectively understand complex analysis within the videos, significantly improving its ability to provide contextual-aware and accurate responses.

\subsection{Long Video Understanding}
In parallel, we detailedly introduce video-specialized models~\cite{he2024ma, song2024moviechat, cheng2024videollama} integrated with LLMs, which have also been widely explored these days to enhance video understanding and reasoning. Here, the most challenging part of current video-LMMs lies in handling long video sequences, mainly due to the limited context length of the LLMs. This limitation compels the models to sparsely sample the video frames in only limited sizes (\textit{e.g.,} typically 8 or 16 frames), potentially missing important spatial and temporal information. To address this, several studies have focused on compressing visual tokens into a more manageable size, proposing aggregation~\cite{maaz2023video, li2025llama} or pooling methods~\cite{lin2023video, xu2024pllava} with advanced vision encoder structures~\cite{zhai2023sigmoid, zhulanguagebind}. In addition, memory-augmented methods~\cite{he2024ma, song2024moviechat} first stored long-term information in a memory bank, then responded to specific queries by loading memory features from the stored buffer. 

On the other hand, among more recent approaches, Li \textit{et al.}~\cite{zhang2024long} have directly extended the LLMs' context length by exploiting RoPE-based frequency interpolation, and Xue \textit{et al.}~\cite{xue2024longvila} have introduced sequence parallelism that can be implemented on multiple GPUs by modifying backend systems. However, we argue that such approaches inherently cannot be free from the fixed context length and provoke intensive memory demands when processing more longer videos. Instead, by focusing on the relevant segments within the entire video, SALOVA can efficiently handle the limited context length, enabling targeted processing of key moments without the need for excessive memory, thereby enhancing performance on longer video sequences.
\section{SceneWalk Dataset}
\label{sec:data}

In this section, we elaborate on how we collected the SceneWalk dataset. The overall pipeline for building the dataset and summarized statistics are illustrated in~\cref{fig:1}. While several video SFT datasets~\cite{maaz2023video, li2024mvbench} are widely used during the instruction tuning stage, they often fail to capture comprehensive details within the scenes. This stems from the nature of instruction-type questions, which provide only partial information, and the brief lengths of both videos and texts in QAs. In contrast, the SceneWalk dataset offers densely captioned video-text pairs that cover long sequence videos in full details, as shown in Fig.~\ref{fig:1}(a). For further detailed data statistics, please see Appendix A.

\subsection{Data Gathering and Processing}
\paragraph{Video Source \& Filtering.} For the long and untrimmed video sources, we primarily focus on three key aspects to build densely captioned video dataset: (\lowercase\expandafter{\romannumeral1}) extensive video length with diverse video source categories, (\lowercase\expandafter{\romannumeral2}) high-quality video contents, 
(\lowercase\expandafter{\romannumeral2}) frequent scene transitions within each video. Accordingly, our data collection is mainly sourced from YouTube, ensuring rich dynamic content that better reflects real-world complexities experienced by global users\textemdash here, because our main goal for video gathering is on complex scene understanding, we exclude low-quality and user-uploaded aesthetic videos (\textit{e.g.,} WebVid, Pixabay, Pexels, and  etc,.) that are rather beneficial for video generation tasks, despite their merits for easy collection. We have collected YouTube urls from~\cite{ju2024miradata} and downloaded the whole video in untrimmed states. Among the total $32$ coarse and diverse video categories YouTube API provided, we selectively curated $10$ categories, excluding categories such as News \& Politics, Classics, and Documentary, due to their static nature, which provides only sparse temporal information in the videos. We further supplemented the dataset with additional Movie \& Drama videos sourced from~\cite{song2024moviechat, ghermi2024short}, totaling $87,867$ video sources with $11.87$ Khrs video duration (avg. $486.5$-seconds).

\paragraph{Segmenting Video into Clips.} Next, for the collected long and untrimmed video sources, we cut the lengthy videos into small segments to densely caption the entire video in next phase. Instead of adopting the bottom-up approach used in the ShareGPT4Video dataset~\cite{chen2024sharegpt4video}, which segments videos into fixed time intervals (2-seconds) in advance and then merges adjacent frames based on their CLIP similarity~\cite{radford2021learning}, we directly employed PysceneDetect~\footnote{We use AdaptiveDetector with default setup in \url{https://github.com/Breakthrough/PySceneDetect}} to segment the videos, dynamically adjusting the threshold based on the raw-level video information to reliably detect scene changes. At the end, the total number of $1.29$M of video segments with $33.11$-seconds average video length is extracted from the original video sources.

\subsection{Captioning and Scoring}
\label{sec:cap_score}

\paragraph{Dense Segment Captioning.} After obtaining the massive video segments, our next goal is to caption each segment with visual details and narrative context to capture the scene-specific explanations, which can enrich scene-level interpretation. To achieve this, we plan to utilize pre-trained LMMs to generate detailed descriptions for the partial video segments. As the captioner, we empirically found that VILA-1.5 (13B)~\cite{lin2024vila} shows competent descriptive quality than other open-sourced models, and used to generate dense captions for each video segment with randomly sampled instructions for detailed descriptions. As a result, we acquire $1.29$M pairs of detailed descriptions corresponding to the video segments, each description with average $137.5$ word length. Please see instruction details and qualitative examples of generated captions in Appendix A.

\paragraph{Scoring Video-Text Correspondence.} Lastly, we score the correspondence between the video segments and the paired dense descriptions, which will later be used as explicit supervision to robustly train our retrieval framework. What we must not overlook here is that the paired video-text relationship is not solely a one-to-one correspondence but is more akin to generalized bipartite matching. That is, within the long and untrimmed video source, each video segment can be connected to other descriptions with additional edges. Therefore, for the $N_{v}$ number of video segments and their paired segments, we can construct a $\{N_{v}\}^{2}$ correspondence matrix between video-text (V2T). To measure each correspondence, we employ LanguageBind~\cite{zhulanguagebind} due to its competitive alignment capabilities across various modalities. In addition, we build another $\{N_{v}\}^{2}$ matrix to provide a doubly robust measure for the correspondence scores among adjacent descriptions (T2T) by comparing similarity within the textual context using the SBERT model~\cite{thakur-2020-AugSBERT}.
% %################################################################################
% Figure
\begin{figure*}[t!]
\centering
\includegraphics[width=1.0\linewidth]{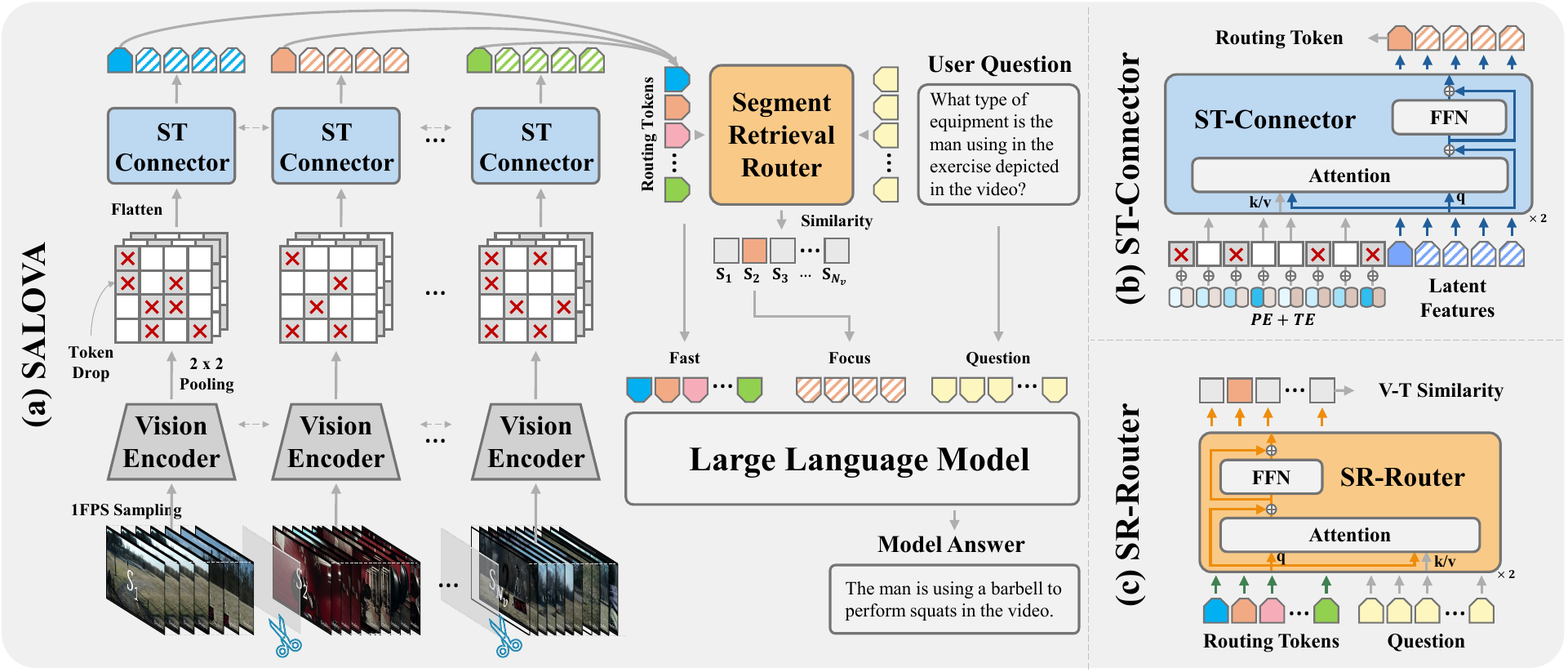}
\vspace*{-0.9cm}
\caption{The network overview of SALOVA. Our framework consists of four structural components: vision encoder, ST-connector, SR-router, and LLMs. Using the FocusFast strategy, our model can concentrate on more detailed local information while maintaining context awareness.}
\vspace*{-0.4cm}
\label{fig:2}
\end{figure*}
% %################################################################################

\section{Segment-Augmented LOng Video Assistant}
\label{sec:method}
\paragraph{Network Overview.}
For a given set of $N_{v}$ video segments sampled at 1 FPS $v{=}\{v_{i}\}^{N_{v}}$, where each segments $v_{i}\in\mathbb{R}^{T_{i}\times H \times W \times C}$ has varying video length (summing up to the total time $T$ of a long and untrimmed video), SALOVA consists of four main architecture components as illustrated in \cref{fig:2}:
\begin{itemize}
    \item Vision Encoder: We use CLIP~\cite{radford2021learning} or SigLIP~\cite{zhai2023sigmoid} to extract visual features, followed by 2x2 average pooling, resulting in $144$ or $196$ visual tokens per frame.
    \item Spatio-Temporal Connector: To handle spatio-temporal features of varying lengths from the vision encoder, we employ the Perceiver Resampler~\cite{alayrac2022flamingo}, which consists of 2-layer Transformer architecture followed by a 2-layer MLP with GELU activation as projector. This resampler embeds each video segment feature into fixed size latent features that are connected to LLMs.
    \item Segment Retrieval Router: For the given textual queries, a retrieval structure (2-layer Transformer) gathers representative information (\textit{i.e.,} routing tokens) from each video segment and then routes the query-relevant video features into the LLMs. Note that the router architecture is trained in an end-to-end manner.
    \item Large Language Model: For LM backbones, We utilize three open-sourced LLMs with varying parameter sizes, LLaMA-3.2 (3B)~\cite{dubey2024llama}, Phi-3.5 (3.8B)~\cite{abdin2024phi} and Qwen-2.5 (7B)~\cite{yang2024qwen2}, all of which are instruction-tuned models that possess QA assistant capabilities.
\end{itemize}
% We will detailedly explain the whole process pipelines for the long and untrimmed video in the following subsections.

\subsection{Long Video Processing and Pipeline}
\subsubsection{Spatio-Temporal Connector}
The first component of our model, Spatio-Temporal Connector, efficiently handles long and variable-length input video segments by extracting each segment's visual semantics in a fixed-size latent vector. As illustrated in \cref{fig:2}(b), we first sample video frames at 1 FPS from each video segment, then visual features are acquired with $2 \times 2$ pooling (thus, $196$ tokens from each frame). After that, the visual features are flattened and fed into the ST-Connector with additional positional and temporal encoding. Here, when the long video is processed, the number of unfolded patch tokens becomes extremely large, leading to exhaustive computations. To address this, we employ a dynamic token drop technique to reduce computational load.

\paragraph{Dynamic Token Drop.}
To effectively manage long video sequences, the token drop has been utilized in video generation tasks~\cite{dehghani2024patch, liu2024sora}. Expanding such approach, in our framework, the dropout rate is dynamically adjusted based on the length of the input sequence $T_i$ in the input visual feature $f_v \sim T_i \times H_p W_p \times d$, which allows for more efficient processing of longer sequences by reducing computational demands, while still preserving dense visual semantics in shorter videos. Additionally, to retain spatio-temporal information from the dropped patches, we add positional embeddings separately along the spatial and temporal axes. This enables more refined extraction of spatio-temporal visual semantics even after reducing the number of tokens.

\subsubsection{Segment Retrieval Router}
Next, the key to conveying the pertinent video information to LLMs is retrieving relevant video segments by querying sentence. To densely cue the similarities between the video and sentence information, we introduce a routing framework, Segment Retrieval Router, which consists of 2-layer Transformer as illustrated in \cref{fig:2}(c). After obtaining the routing tokens $R{=}\{R_{i}\}^{N_{v}}\in\mathbb{R}^{N_{v}\times D}$ from entire video segments, we aggregate them and feed into the SR-Router as queries. For the given sentence, we employ the same text encoder used for the vision encoder and project it into the shared embedding space to obtain sentence features $S\in\mathbb{R}^{N_{t}\times D}$, where $N_{t}$ indicates textual length.

Using the cross attention mechanism (q: $R$; k/v: $S$), we can estimate similarity scores between the video segments and given sentence queries (\textit{i.e.,} V-T similarity). The scores enable the SR-Router to prioritize and select the most relevant video segments that align with the sentence query.

\paragraph{Retrieval Objective.}
To seamlessly train the SR-Router with the mainstream flows of SALOVA in an end-to-end manner, we have designed a similarity loss function $\mathcal{L}_{\text{sim}}$ that minimizes the distance between the high-dimensional embeddings of the video segments and sentence queries. Here, we use the correspondence scores (aforementioned in \cref{sec:cap_score}) as a retrieval supervision signal $y_{i}$, after applying one-hot encoding. We incorporate a simple margin-based loss, commonly used in contrastive learning settings~\cite{li2024momentdiff}, which enables the model to learn off-diagonal relaxation in the correspondence matrices between video segments and sentences. As mentioned earlier, the relationship between paired videos and sentences is closer to generalized bipartite matching than to one-to-one matching, so relaxation learning helps to accommodate the inherent complexity in aligning correspondence. In conclusion, with the binary cross-entropy loss and the score margin loss, we can formulate the similarity loss as follows:
%%%%%%%%%%%%%%%%%%%%%%%%%%%%%%%%%%%%%%%%%%%%%%%%%%%
\begin{equation}
\label{eqn:sim}
    \mathcal{L}_{\text{sim}}=\underbrace{\mathcal{L}_{\text{bce}}{(y_{i},s_{i})}_{i=1}^{N_{v}}}_{\text{point-wise CE}}+\underbrace{\textstyle\frac{1}{N_{s}}\textstyle\sum_{j} \max\left(0, \delta-(s^{p}_{j}{-}s^{n}_{j})\right)}_{\text{score margin loss}},
\end{equation}
%%%%%%%%%%%%%%%%%%%%%%%%%%%%%%%%%%%%%%%%%%%%%%%%%%%
where $s^{p}_{j}$ and $s^{n}_{j}$ indicate randomly sampled scores from positive and negative pairs, respectively, and $\delta$ denotes the margin parameter (set as $0.2$). Note that the similarity loss is trained in conjunction with the auto-regressive loss $\mathcal{L}_{\text{ar}}$ from subsequent LLMs in an end-to-end manner.

\subsubsection{FocusFast Pathways: Integration to LLMs}
Using the routing tokens, we can calculate the similarities of each video segment for the given query. Leveraging the similarities, SALOVA efficiently retrieves the specific video features that exhibit the highest relevance score to the textual query, where the indexed video features are then directly integrated into the LLM architecture. Here, extending the SlowFast pathways concept~\cite{feichtenhofer2019slowfast}, we present the FocusFast mechanism to effectively manage the processing pathway for the retrieved video segments: (\lowercase\expandafter{\romannumeral1}) Focus pathway concatenates the top-K most pertinent features to construct a comprehensive video representation, capturing local details across retrieved segments and enabling detailed interactions with textual queries to enhance handling complex video information. (\lowercase\expandafter{\romannumeral2}) Fast pathway focuses on the more broader-level context by employing segment-wide routing tokens as the condensed global representation. It effectively contains dynamic spatio-temporal changes throughout the video stream, thereby allowing SALOVA to understand the overall video content and scene-level continuity awareness.

Once the most pertinent features are retrieved, they are delivered to the LM backbone for the final processing as in \cref{fig:2}(a), integrating video-specific details into the models' responses. By effectively handling long and untrimmed videos with the proposed retrieval and routing mechanism, SALOVA can maintain the flow of salient information without the processing overhead for less related data, thus generating more context-aware responses.

\subsection{Training Strategies}
\label{subsec:training}
The current training strategies for LMMs predominantly consist of two-step training: (\lowercase\expandafter{\romannumeral1}) cross-modal alignment and (\lowercase\expandafter{\romannumeral2}) visual instruction tuning. Recently, Li \textit{et al.}~\cite{li2024llava} have emphasized the importance of high-quality knowledge learning between the two training stages (thus stage 1.5), pointing out that the models cannot enoughly learn necessary knowledge during the alignment with the low-qualitative web-scale image-text data. As the similar approach of using rephrased descriptions for additional knowledge learning~\cite{li2024llava}, we employ the newly collected SceneWalk dataset as the parametric knowledge injection step, which enables the SALOVA to learn detailed spatial and temporal representation from the long sequence video data before the instruction tuning. Accordingly, our training recipes and data configuration can be divided into three steps as follow (Please see training details in Appendix B):

\paragraph{Stage 1: Cross-modality Alignment.} For the initial step in modality alignment, we utilize $790$K image/video-text paired dataset: (\lowercase\expandafter{\romannumeral1}) $558$K image-text pairs from the CC3M dataset~\cite{sharma2018conceptual}, filtered by LLaVA~\cite{liu2023visual} and (\lowercase\expandafter{\romannumeral2}) video-text pairs sampled from the WebVid $2.5$M subset~\cite{bain2021frozen}. We freeze vision encoder and LLMs during the training, and mainly focus on optimizing the connector and router to map the visual information into the textual space. 

\paragraph{Stage 1.5: Long Video Knowledge Injection.} As an intermediate training step, we use the SceneWalk dataset to train the SALOVA, unfreezing all trainable parameters except for the vision encoder. During training, we input the long and untrimmed video instances and follow the processing pipeline shown in \cref{fig:2}. By training the model with densely captioned video-description pairs, it acquires high-quality parametric knowledge of both spatial and temporal information. In addition, through the aforementioned retrieval process, the model learns to target video segments that are mostly relevant to the video description.

\paragraph{Stage 2: Video Instruction Tuning.} To possess QA capabilities in SALOVA, we use extensive video instruction-tuning data as the final training step. The instruction data are mainly sourced from four different datasets: LLaVA-Video-178K~\cite{zhang2024video}, NeXT-QA~\cite{xiao2021next}, ActivityNetQA~\cite{yu2019activitynet}, and PerceptionTest~\cite{patraucean2024perception}\textemdash comprising a total of $1.4$M video-instruction QA data, including caption entries, open-ended QA, and multiple-choice QA. Note that we train all the network parameters during this stage and auto-regressively update the instruction-following assistant's response for the next word prediction.   
\section{Experiments}

\subsection{Experimental Details}
\paragraph{Implementation.} For the vision and text encoders of the SR-Router, we utilize CLIP~\cite{radford2021learning} for the small-size models and SigLIP~\cite{zhai2023sigmoid} for the frontier model, with resolution sizes of $336$ and $384$, respectively. We employ a 2-layer transformer with a head size of 2 for the ST-Connector, which has a latent dimension of $256$. The token drop mechanism is dynamically applied based on video length, with varying maximum drop rates at each training stage\textemdash Stage 1 has no token drop, Stage 1.5 allows up to $0.7$, and Stage 2 allows up to $0.4$ For the configuration of SR-Router, we set 2-layer of transformers with a single head, and top-K number is set to $5$ during the stage 2. Following the recent works~\cite{liu2023improved, li2024llavanext}, the projector layer consists of 2-layer MLP with GELU. We employ three LLMs: (\lowercase\expandafter{\romannumeral1}) 3B: Llama-3.2-3B~\cite{dubey2024llama}, (\lowercase\expandafter{\romannumeral2}) 3.8B: Phi-3.5~\cite{abdin2024phi}, and (\lowercase\expandafter{\romannumeral3}) 7B: Qwen2.5-7B~\cite{yang2024qwen2}.

\paragraph{Training Details.} For the each training stage, we train SALOVA for $1$ epoch with 1 node of 8 A100 GPUs. The total training hours for 3B (3.8B) and 7B models roughly take 5 and 7 days, respectively. We employ FlashAttention-2~\cite{dao2023flashattention}, gradient checkpointing~\cite{chen2016training}, and ZeRO-2~\cite{rajbhandari2020zero} to minimize the memory footprint associated with model components (\textit{i.e.,} gradient, activation, and optimizer states). Additionally, we fine-tune the trainable parameters at each step without employing LoRA~\cite{hu2021lora}. For the extended training config, we have attached the details in Appendix C.

\paragraph{Evaluation Benchmarks.} We evaluate our model using two types of video analysis benchmarks—long video understanding and general video understanding, categorized based on the video length. For the long video benchmark, we primarily utilize Video-MME~\cite{fu2024video} and LongVideoBench~\cite{wu2024longvideobench}, both of benchmarks includes videos up to two hours long duration. As the general video analysis evaluations, we employ various benchmarks such as ActivityNetQA~\cite{yu2019activitynet}, VideoChatGPT~\cite{maaz2023video}, and MVBench~\cite{li2024mvbench}. Note that the same pipeline is used to obtain video segments for each benchmark, and all benchmarks are sampled at 1 FPS without token drop during inference. As comparison baselines, considering academic budget constraints, we evaluate against models that have similar parameter size.

%%%%%%%%%%%%%%%%%%%%%%%%%%%%%%%%%%%%%%%%%%%%%%%%%%%%%%%%%%%%%%%%%%%%%%%%%%%%%%%%%%%%
\begin{table}[t!]
\centering
\resizebox{1.0\linewidth}{!}{
\begin{tabular}{lcccccc}
\Xhline{3\arrayrulewidth}\rule{0pt}{9pt}
                        &                          & \multicolumn{4}{c}{Video-MME}                                                                                         & \multicolumn{1}{l}{LVBench} \\ \cmidrule(lr){3-6} \cmidrule(lr){7-7}
\multirow{-2}{*}{Model} & \multirow{-2}{*}{\#param} & Short                       & Medium                      & Long                        & Overall                     & Acc. (val)                        \\ \hline
\multicolumn{7}{l}{\cellcolor[HTML]{EFEFEF}Proprietary LMMs}                                                                                                                                            \\
GPT-4V~\cite{gpt4v}                  & n/a                      & 70.5                        & 55.8                        & 53.5                        & 59.9                        & -                          \\
GPT-4o~\cite{gpt4o}                  & n/a                      & 80.0                        & 70.3                        & 65.3                        & 71.9                        & 66.7                       \\
Gemini 1.5 Pro~\cite{reid2024gemini}         & n/a                      & 81.7                        & 74.3                        & 67.4                        & 75.0                        & 64.0                       \\ \cdashline{1-7}
\multicolumn{7}{l}{\cellcolor[HTML]{EFEFEF}Open-sourced LMMs}                                                                                                                                           \\
ST-LLM~\cite{liu2025st}                  & 7B                       & 45.7                        & 36.8                        & 31.3                        & 37.9                        & -                          \\
VideoChat2~\cite{li2024mvbench}      & 7B                       & 48.3                        & 37.0                        & 33.2                        & 39.5                        & 39.3                       \\
ShareGPT4Video~\cite{chen2024sharegpt4video}          & 8B                       & 48.3                        & 36.3                        & 35.0                        & 39.9                        & 39.7                       \\
Video-LLaVA~\cite{lin2023video}             & 7B                       & 45.3                        & 38.0                        & 36.2                        & 39.9                        & 39.1                       \\
Chat-UniVi-V1.5~\cite{jin2024chat}         & 7B                       & 45.7                        & 40.3                        & 35.8                        & 40.6                        & -                          \\
Qwen-VL-Chat~\cite{bai2023qwen}           & 7B                       & 46.9                        & 38.7                        & 37.8                        & 41.1                        & -                          \\
ShareGemini~\cite{sharegemini}             & 7B                       & 49.1                        & 41.3                        & 39.1                        & 43.2                        & -                          \\
SliME~\cite{zhang2024beyond}                   & 8B                       & 53.3                        & 42.7                        & 39.8                        & 45.3                        & -                          \\
PLLaVA~\cite{xu2024pllava}                  & 7B                       & -                           & -                       & -                           & -                           & 40.2                           \\
VideoLLaMA2~\cite{cheng2024videollama}             & 8B                       & 56.0                        & 45.4                        & 42.1                        & 47.9                        & -                          \\ 
LongVA~\cite{zhang2024long}             & 7B                       & \textbf{61.1}                        & \textbf{50.4}                        & \underline{46.2}                        & \underline{52.6}                        & -                          \\ \cdashline{1-7}
\multicolumn{7}{l}{\cellcolor[HTML]{EFEFEF}Ours}                                                                                                                                                        \\
SALOVA-3B            & 3B                       & 48.3	                     & 46.3                        & 41.1                        & 45.3                        & 41.4                         \\
SALOVA-3.8B              & 3.8B                     & 47.1	                     & \underline{48.8}	                       & 44.1                        & 46.7                        & \underline{41.6}                           \\
SALOVA-7B             & 7B                       & \underline{59.4}	                     & \textbf{50.4}	                       & \textbf{49.4}	                     & \textbf{53.1}                        & \textbf{44.6}                       \\
\Xhline{3\arrayrulewidth}\rule{0pt}{9pt}
\end{tabular}
}
\vspace{-0.8cm}
\caption{Detailed results for the Video-MME benchmark (w/o subtitles) and LongVideoBench. The best results are highlighted in \textbf{bold} and the runner-up results are \underline{underlined}.}
\vspace{-0.4cm}
\label{table:longvideo}
\end{table}

%%%%%%%%%%%%%%%%%%%%%%%%%%%%%%%%%%%%%%%%%%%%%%%%%%%%%%%%%%%%%%%%%%%%%%%%%%%%%%%%%%%%

\subsection{Experimental Results}

\paragraph{Results on Long Video Understanding.}
Video-MME~\cite{fu2024video} evaluates LMMs with a focus on video analysis across a variety of video types and durations. We primarily compare the benchmark results in settings without subtitles, relying solely on video frames. Therefore it can assess the LMMs' visual comprehension capabilities rigorously, based purely on visual content. Also, LongVideoBench~\cite{wu2024longvideobench} is designed to assess LMMs' understanding of long-duration videos up to two hours. It includes a diverse collection of videos, challenging the models' ability to process and interpret extensive visual and contextual information across a variety of themes. As shown in \cref{table:longvideo}, our model shows competent video understanding performance across all video length distributions in Video-MME and lengthy video instances in LongVideoBench. Notably, we highlight that SALOVA achieved significant performance in the medium (average 562.7 seconds) and long (average 2385.8 seconds) length categories in Video-MME benchmark, even with more smaller size of backbone LM parameters compared with the baseline models. 

%%%%%%%%%%%%%%%%%%%%%%%%%%%%%%%%%%%%%%%%%%%%%%%%%%%%%%%%%%%%%%%%%%%%%%%%%%%%%%%%%%%%
\begin{table}[t!]
\centering
\resizebox{1.0\linewidth}{!}{
\begin{tabular}{lccccc}
\Xhline{3\arrayrulewidth}\rule{0pt}{9pt}
                        &                          & \multicolumn{2}{c}{ActivityNetQA}                & VideoChatGPT & MVBench \\ \cmidrule(lr){3-4} \cmidrule(lr){5-5} \cmidrule(lr){6-6}
\multirow{-2}{*}{Model} & \multirow{-2}{*}{\#param} & \multicolumn{2}{c}{test (acc/score)}                         & test (acc)         & test (acc)   \\ \hline
\multicolumn{6}{l}{\cellcolor[HTML]{EFEFEF}Proprietary LMMs}                                                                   \\
GPT-4V~\cite{gpt4v}                  & n/a                      & 57.0                     & \multicolumn{1}{c}{-} & 4.06         & 43.5    \\
GPT-4o~\cite{gpt4o}                  & n/a                      & 61.9                     & \multicolumn{1}{c}{-} & -            & -       \\
Gemini 1.5 Pro~\cite{reid2024gemini}          & n/a                      & 57.5                     & \multicolumn{1}{c}{-} & -            & -       \\ \cdashline{1-6}
\multicolumn{6}{l}{\cellcolor[HTML]{EFEFEF}Open-sourced LMMs}                                                                  \\
VideoLLaMA~\cite{zhang2023video}              & 7B                       & \multicolumn{1}{r}{12.4} & 1.1                   & 2.16         & 34.1    \\
VideoChatGPT~\cite{maaz2023video}            & 7B                       & \multicolumn{1}{r}{35.2} & 2.7                   & 2.42         & 32.7    \\
MovieChat~\cite{song2024moviechat}               & 7B                       & \multicolumn{1}{r}{45.7} & -                     & 2.67         & -       \\
Chat-UniVi~\cite{jin2024chat}              & 7B                       & \multicolumn{1}{r}{46.1} & 3.2                   & 2.99         & -       \\
LLaMA-VID~\cite{li2025llama}               & 7B                       & \multicolumn{1}{r}{47.4} & 3.3                   & 2.89         & 41.3    \\
VideoChat2~\cite{li2024mvbench}              & 7B                       & \multicolumn{1}{r}{49.1} & 3.3                   & 2.98         & 51.1    \\
VideoLLaMA2~\cite{cheng2024videollama}             & 8B                       & \multicolumn{1}{r}{50.2} & 3.3                   & \textbf{3.13}         & \textbf{54.6}    \\ \cdashline{1-6}
\multicolumn{6}{l}{\cellcolor[HTML]{EFEFEF}Ours}                                                                               \\
SALOVA-3B            & 3B                       & \multicolumn{1}{r}{\underline{52.6}} & \underline{3.4}         & \underline{3.08}             & 51.7        \\
SALOVA-3.8B              & 3.8B                     & \multicolumn{1}{r}{51.1} & \textbf{3.5}         & 2.83             & 46.4        \\
SALOVA-7B             & 7B                       & \multicolumn{1}{r}{\textbf{53.6}} & \textbf{3.5}         & \textbf{3.13}         & \underline{53.5}         \\
\Xhline{3\arrayrulewidth}\rule{0pt}{9pt}
\end{tabular}
}
\vspace{-0.8cm}
\caption{Comparison results for generic video understanding benchmarks. The best results are highlighted in \textbf{bold} and the runner-up results are \underline{underlined}.}
\vspace{-0.4cm}
\label{table:general_table}
\end{table}
%%%%%%%%%%%%%%%%%%%%%%%%%%%%%%%%%%%%%%%%%%%%%%%%%%%%%%%%%%%%%%%%%%%%%%%%%%%%%%%%%%%%

Such performance gains in long video instances are attributed to our model's dynamic capability to retrieve and process only the relevant video segments, enabling it to handle lengthy video content efficiently without being constrained by the limited context length. Especially, the routing mechanism in SALOVA strategically prioritizes video segments that are likely to contain crucial visual and contextual cues relevant to the query. This selective routing mechanism reduces the computational load and minimizes the information loss that commonly occurs in current video-LMMs trying to process extensive video data in entirety.

\paragraph{Results on General Video Understanding.}
For general video understanding benchmarks such as ActivityNetQA~\cite{yu2019activitynet}, VideoChatGPT~\cite{maaz2023video}, and MVBench~\cite{li2024mvbench}, SALOVA was evaluated across various video types to assess its general video understanding capabilities. As shown in \cref{table:general_table}, SALOVA demonstrated competent performance, comparable to existing video-LMMs, especially in dynamic and shorter video sequences. On ActivityNetQA, the model effectively utilized its segment retrieval strategy to provide focused and contextually appropriate responses, which helped maintain accuracy. This approach was similarly effective in the multi-modal settings of VideoChatGPT and MVBench, where SALOVA showed consistent performance in handling dialogues and visual cues. These outcomes highlight SALOVA's capability to process general video content efficiently through its dynamic routing mechanism, offering a reliable solution that balances computational resources with output quality.

%################################################################################
\begin{table}[t!]
\centering
\resizebox{1.0\linewidth}{!}{
\begin{tabular}{ccccc}
\Xhline{3\arrayrulewidth}\rule{0pt}{9pt}
\multirow{2}{*}{Ablation} & \multicolumn{4}{c}{Video-MME}                                \\ \cmidrule(lr){2-5}
                          & Short: $\leq$2m    & Mid: 4-15m    & Long: 30-60m   & Overall     \\ \hline
\rowcolor[HTML]{EFEFEF} 
frm sample                 & \multicolumn{4}{l}{\cellcolor[HTML]{EFEFEF}: Frame sampling rate (w/o SR-Router)}                     \\
8 frm                     & 48.3	      & 42.0	      & 37.2           & 42.5             \\
16 frm                    & 50.0	      & 42.8	      & 38.0           & 43.6             \\
1 FPS                     & 48.3          & 46.3	      & 41.1	       & 45.3        \\ \cdashline{1-5}
\rowcolor[HTML]{EFEFEF} 
1 / 1.5 / 2                & \multicolumn{4}{l}{\cellcolor[HTML]{EFEFEF}: Train stage - Long video knowledge injection} \\
\Checkmark~~\ding{55}~~\Checkmark  & 45.6	       & 43.7	      & 40.2	       & 43.6        \\
\Checkmark~~\Checkmark~~\Checkmark & 48.3          & 46.3	      & 41.1	       & 45.3        \\ \cdashline{1-5}
\rowcolor[HTML]{EFEFEF} 
FocusFast                   & \multicolumn{4}{l}{\cellcolor[HTML]{EFEFEF}: Local-global video representation}        \\
\ding{55}                 & 36.4          & 38.6	      & 35.6           & 36.9        \\
\Checkmark                & 48.3          & 46.3	      & 41.1	       & 45.3        \\ 
\Xhline{3\arrayrulewidth}\rule{0pt}{9pt}
\end{tabular}
}
\vspace{-0.8cm}
\caption{Ablation studies on SALOVA configuration. We utilize SALOVA-3B for the resource efficiency.}
\vspace{-0.3cm}
\label{table:ablation}
\end{table}

%################################################################################

\subsection{Additional Analyses on SALOVA}
\paragraph{Ablation Study.} We conduct ablation studies on three components as follows: (\lowercase\expandafter{\romannumeral1}) different video frame sampling strategies, (\lowercase\expandafter{\romannumeral2}) intermediate training stage for long video knowledge injection, and (\lowercase\expandafter{\romannumeral3}) the FocusFast mechanism to understand branched local-global representation in videos. 

As in Table \ref{table:ablation}, We first observe that using more frames with SR-Router significantly enhances performance, particularly in long-form videos. This aligns with our key contributions on managing long videos through an effective routing mechanism, suggesting that retrieving more frames can provide richer spatio-temporal information and improve the model's responses without losing contextual coherence. Additionally, we compare with a baseline trained with stage 1-2 (skipping stage 1.5). Here, we highlight the effectiveness of the SceneWalk dataset as an intermediate training step to enhance parametric knowledge for the long video analysis by allowing the model to learn from high-quality and densely captioned scene-level information, which is crucial for adapting to various lengths and contexts. Lastly, we conduct an analysis on the FocusFast method and demonstrate its efficacy in analyzing not only local details from relevant video segments but also in understanding the global video context through the simultaneous use of routing tokens, thereby facilitating a more comprehensive understanding of video content.

\paragraph{Analysis of Retrieving Segments.} By retrieving relevant video segments for the given queries, SALOVA can effectively target salient information in the long video and retain long context information. To further demonstrate the model's targeting capabilities beyond numerical performance in long video analysis, we explore our model's application in the Visual Needle-In-A-Haystack (V-NIAH) task~\cite{zhang2024long}, which extends the Needle-in-a-Haystack (NIAH) evaluation for LLMs to a vision-level benchmark. This task is particularly challenging as it requires models to not only detect but also precisely retrieve the sparse yet crucial visual cues scattered across lengthy videos. 

\begin{figure}[t!]
\centering
\includegraphics[width=0.99\linewidth]{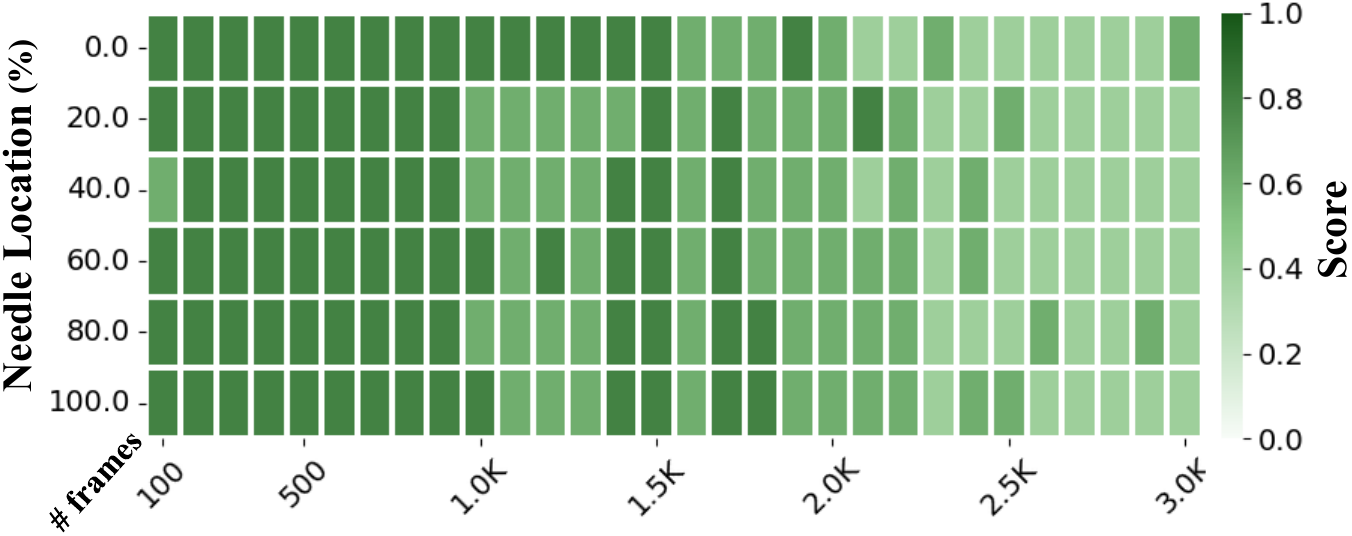}
\vspace*{-0.5cm}
\begin{flushleft}
    \hspace{1.8cm}{(a) SALOVA-3B (16 frm sample)}
\end{flushleft}	
\vspace*{-0.3cm}
\includegraphics[width=0.99\linewidth]{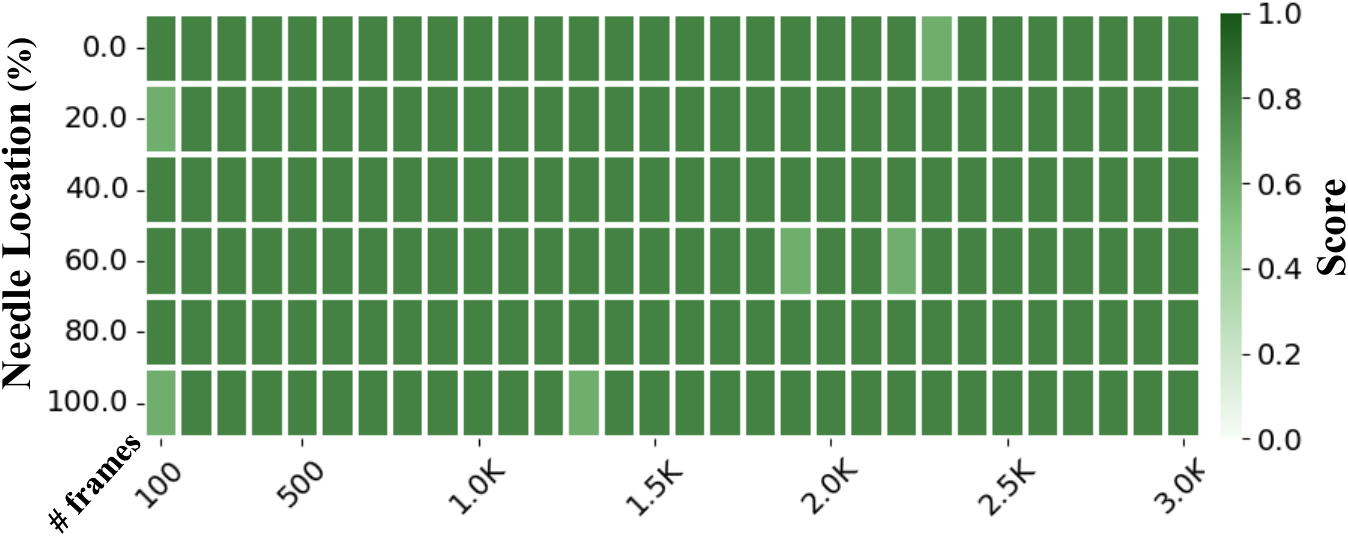}
\vspace*{-0.5cm}
\begin{flushleft}
    \hspace{1.9cm}{(b) SALOVA-3B (1 FPS sample)}
\end{flushleft}	
\vspace*{-0.6cm}
\caption{Comparison results of V-NIAH. The x/y-axis indicates the total video frames and the location of needle image within the video, respectively.}
\label{fig:3}
\vspace{-0.4cm}
\end{figure}

As in \cref{fig:3}, we compare SALOVA against a baseline trained on sparsely sampled frames (16 frm w/o SR-Router). Our framework effectively identifies and extracts relevant video segments from densely packed content, even when handling long context lengths. These results highlight SALOVA's robustness in managing complex, long-form videos, maintaining contextual continuity by strategically focusing on relevant segments to user queries.

 %%%%%%%%%%%%%%%%%%%%%%%%%%%%%%%%%%%%%%%%%%%%%%%%%%%%%%%%%%%%%%%%%%%%
%%%%%%%%%%%%%%%%%%%%%%%%%%%%%%%%%%%%%%%%%%%%%%%%%%%%%%%%%%%%%%%%%%%%
\vspace{-0.5mm}
\section{Discussion and Conclusion}
\vspace{-0.5mm}
\paragraph{Discussion.} Despite SALOVA's competence in handling extended video sequences, it is important to recognize scenarios where its complex architecture may not be necessary. Specifically, for shorter videos where sparse sampling suffices to capture essential spatio-temporal information, simpler models could potentially outperform the efficiency of SALOVA without necessitating its extensive processing capabilities. This suggests a future avenue for integrating a hybrid approach based on our framework by dynamically adjusting the complexity of the retrieval and processing mechanisms based on the video length and content density.

\paragraph{Conclusion.} In this paper, we introduce SALOVA, a novel framework designed to enhance the comprehension of long and untrimmed video by leveraging a retrieval-driven approach with new densely captioned dataset, the SceneWalk dataset. SALOVA strategically targets and processes only the relevant video segments, effectively addressing the structural limitations of current Video-LMMs with its Spatio-Temporal Connector and Segment Retrieval Router. Through extensive evaluation on various benchmarks, SALOVA exhibits its robust performance in interpreting complex video content, enhancing efficiency, and improving the understanding of extended videos.

\newpage
\section*{Acknowledgments}
This work was partially supported by two funds: IITP grant funded by the Korea government (MSIT) (RS-2022-II220984) and IITP grant funded by the Korea government (MSIT) (No.2020-0-00004, Development of Previsional Intelligence based on Long-Term Visual Memory Network)
{
    \small
    \bibliographystyle{ieeenat_fullname}
    \bibliography{main}
}

\clearpage
\setcounter{page}{1}
\maketitlesupplementary

\appendix
\section{Details of SceneWalk Dataset}
\label{sec:rationale}

\subsection{Detailed Data Statistics}

We provide a comprehensive analysis of the proposed SceneWalk dataset, focusing on detailed data statistics, including video duration, categorical distribution, and segment-level descriptions. The information emphasizes the versatility and diversity of the dataset, ensuring its applicability for training our video-LLM.

\paragraph{Dataset Composition.} The SceneWalk dataset comprises 87,867 long-form video sources, spanning a total of 11.87 Khrs (average video duration: 486.5 seconds). The video sources are collected from a curated selection of 10 diverse categories as in 
\cref{fig:1} sourced primarily from YouTube, with additional contributions from Movie \& Drama datasets~\cite{song2024moviechat, ghermi2024short}. This ensures a wide range of real-world scenarios, avoiding static categories.

% %################################################################################
% Figure
\begin{figure}[h!]
\centering
\includegraphics[width=1.0\linewidth]{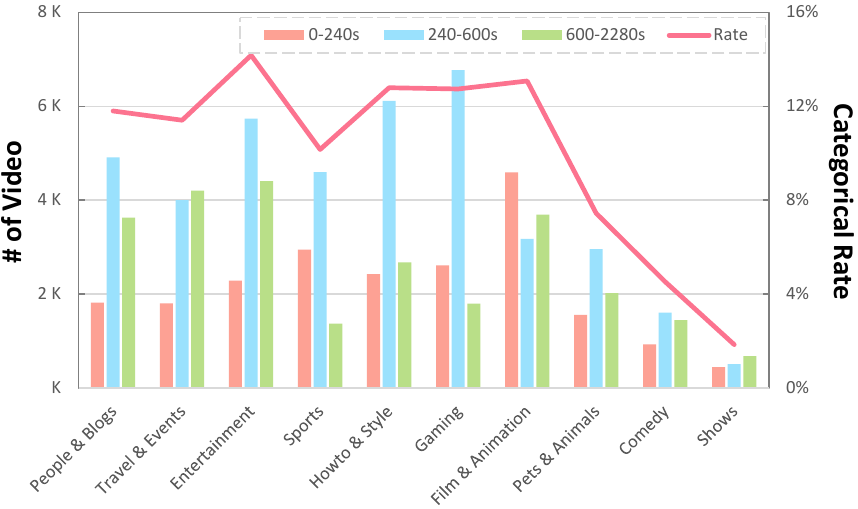}
\vspace*{-0.5cm}
\caption{Detailed video duration range statistics for each video category in the SceneWalk dataset.}
\label{fig:4}
\end{figure}
% %################################################################################

\paragraph{Video Duration Distribution.} The collected videos can be split into three distinct duration ranges to analyze temporal diversity: (\lowercase\expandafter{\romannumeral1}) 0–240 seconds (short): This range constitutes about 24.4\% of all segments, (\lowercase\expandafter{\romannumeral2}) 240–600 seconds (long): This intermediate range accounts for the largest proportion, approximately 46.1\% of the dataset, and (\lowercase\expandafter{\romannumeral3}) 600–2280 seconds (extreme-long): The longest duration range comprises around 29.5\% of the dataset. The distribution of video durations for each video category is illustrated in \cref{fig:1}(b) (outer circle), and more detailed duration distributions can be found in \cref{fig:4}.

\subsection{Pipeline for Dense Caption}
\paragraph{Splitting into Video Segments} To divide untrimmed and long video sources into a massive $1.29$M video segments, we directly utilize PySceneDetect with the AdaptiveDetector using the default adaptive threshold ($3.0$), which compares the difference in content between adjacent frames similar using a rolling average of adjacent frame changes. This can help mitigate false detections in situations such as fast camera motions.

\paragraph{Instructions of Dense Segment Captioning.}
To generate detailed descriptions for each video segment obtained from the above process, we mainly use a pre-trained LMM (VILA-1.5-13B~\cite{lin2024vila}). Below \cref{table:instruction} includes the instructions for generating those captions. We randomly select one from the list and use it as a query for the model.

%%%%%%%%%%%%%%%%%%%%%%%%%%%%%%%%%%%%%%%%%%%%%
%%%%%%%%%%%%%%%%%%%%%%%%%%%%%%%%%%%%%%%%%%%%%%%%%%%%%%%%%%%%%%%%%
\begin{table}[h!]
\centering
\begin{minipage}{0.99\columnwidth}\vspace{0mm}    
\centering
\begin{tcolorbox} 
    \small
\raggedright    
{\color[HTML]{3531FF} \textbf{Video Caption Generation Instruction:}}

\begin{itemize}[leftmargin=3.5mm]
\setlength{\itemsep}{2pt}
\item ``Provide a detailed description of both the visual content and the storyline depicted in the video.''
\item ``Thoroughly describe the scenes, actions, and characters featured in the video''
\item ``Elaborate on the visual and narrative elements of the video in detail.''
\item ``Describe what is happening in the video in detail, including both the visual details and the narrative context.''
\item ``Describe every element of the video, from visual details to the unfolding narrative, including how each aspect interacts to enhance the storytelling.''
\item ``Offer a granular analysis of the video, detailing the scenes, character actions, and dialogue, alongside any symbolic visual elements that add depth to the story.''
\item ``Narrate the unfolding events in the video with attention to both the visual composition and the plot, describing how each scene visually portrays the narrative tensions or themes.''
\end{itemize}

\end{tcolorbox}
\vspace{-0.3cm}
\caption{The list of instructions for detailed description for each video segment.}
    \label{table:instruction}
\end{minipage}
\end{table}
%%%%%%%%%%%%%%%%%%%%%%%%%%%%%%%%%%%%%%%%%%%%%%%%%%%%%%%%%%%%%%%%%
%%%%%%%%%%%%%%%%%%%%%%%%%%%%%%%%%%%%%%%%%%%%%

\begin{figure*}[t!]
\centering
\includegraphics[width=1.0\linewidth]{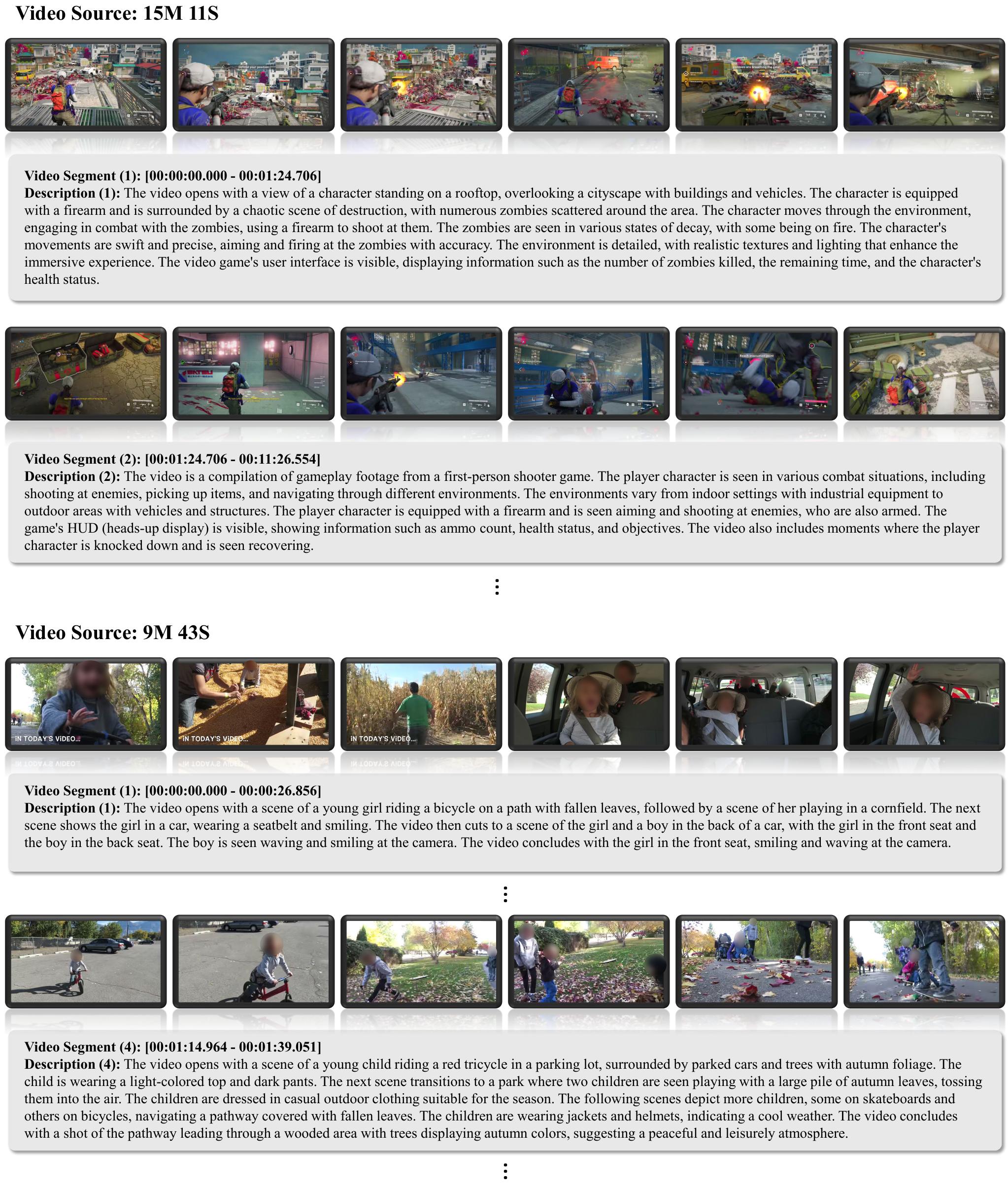}
\vspace*{-0.8cm}
\caption{Examples of the SceneWalk dataset (\lowercase\expandafter{\romannumeral1}).}
\vspace*{-0.5cm}
\label{fig:6}
\end{figure*}

\begin{figure*}[t!]
\centering
\includegraphics[width=1.0\linewidth]{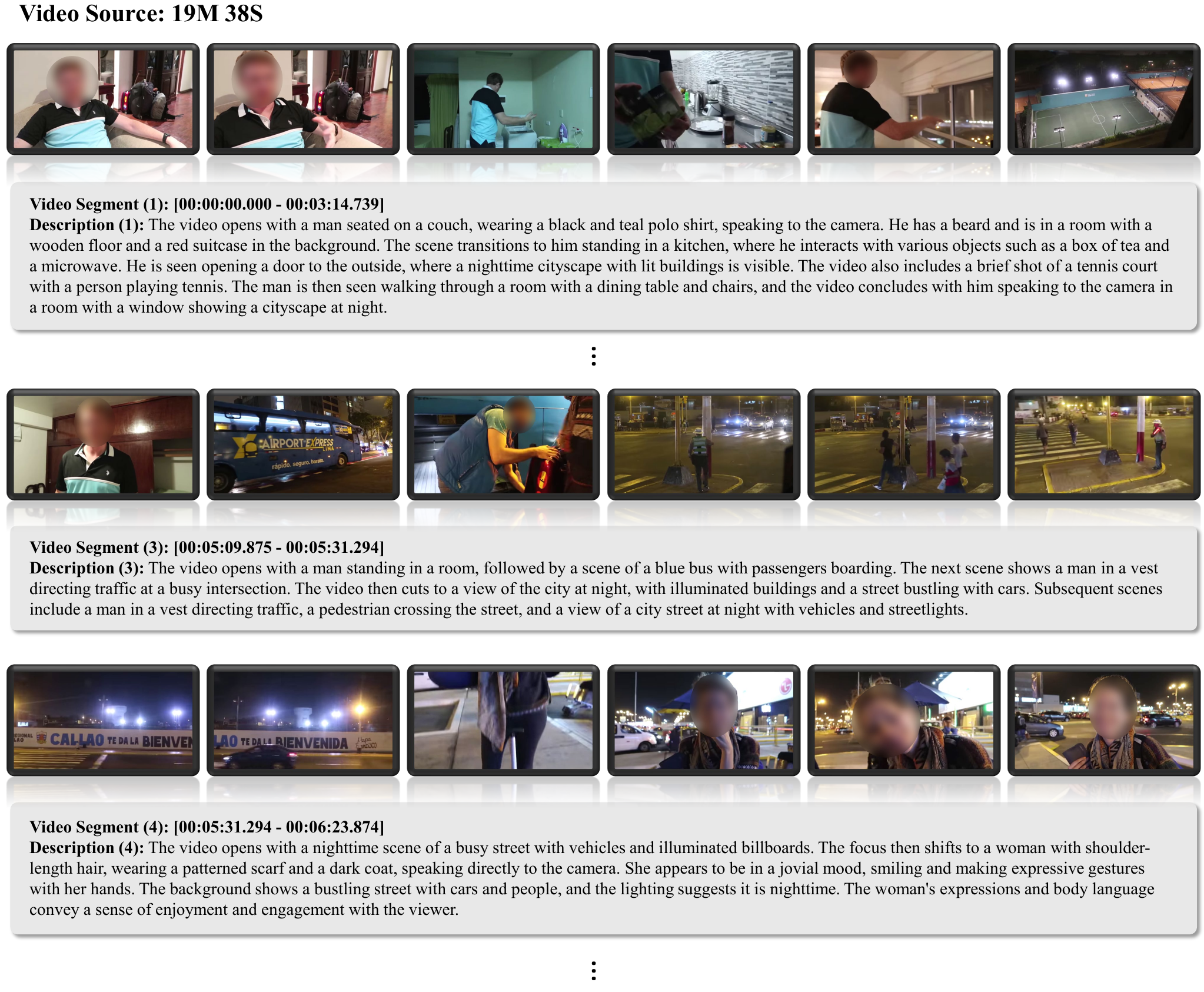}
\vspace*{-0.8cm}
\caption{Examples of the SceneWalk dataset (\lowercase\expandafter{\romannumeral2}).}
\vspace*{-0.5cm}
\label{fig:7}
\end{figure*}

\paragraph{Captioning and Scoring.} Each segment is densely captioned, generating highly detailed textual descriptions that average $137.5$ words per segment. Please see the densely captioned video examples in \cref{fig:6} and \cref{fig:7}. To ensure alignment quality, a generalized bipartite matching framework is employed: (\lowercase\expandafter{\romannumeral1}) Video-to-Text (V2T) Correspondence: A matrix evaluates the alignment between video segments and their paired captions using LanguageBind~\cite{zhulanguagebind}, and (\lowercase\expandafter{\romannumeral2}) Text-to-Text (T2T) Context Similarity: The textual coherence among adjacent captions is assessed using SBERT~\cite{thakur-2020-AugSBERT}, enhancing overall alignment robustness.

\paragraph{Supervision from Correspondence Scores.}
To derive the supervision signal $y_{i}$ for training, we leverage the correspondence scores $S_{\text{V2T}}$ (Video-to-Text) and $S_{\text{T2T}}$ (Text-to-Text), as discussed in \cref{sec:method}. For each correspondence score matrix, we apply thresholding to extract meaningful relationships. Specifically, we define thresholds $\tau_{\text{V2T}}$ and $\tau_{\text{T2T}}$ for the two matrices, and elements with scores exceeding these thresholds are treated as positive correspondences (th: 0.18 ($\tau_{\text{V2T}}$) and 0.8 ($\tau_{\text{T2T}}$), respectively). These positive correspondences are then one-hot encoded to form binary matrices $Y_{\text{V2T}}$ and $Y_{\text{T2T}}$, where each element indicates whether a specific correspondence is valid. Finally, we compute the union of these binary matrices to produce the final supervision signal (\textit{e.g.,} $y_{i} {=} Y_{\text{V2T}} \cup Y_{\text{T2T}}$). The union operation ensures that any correspondence deemed valid by either of the two modalities contributes to the final supervision. This approach captures both the multi-modal alignment (Video-to-Text) and intra-modal coherence (Text-to-Text), providing a robust supervision signal for the retrieval task.

The resulting $y_{i}$ is then incorporated into the similarity loss function $\mathcal{L}_{\text{sim}}$ as described in \cref{eqn:sim}, ensuring that the model effectively learns the nuanced relationships between video segments and their corresponding textual descriptions. By combining $S_{\text{V2T}}$ and $S_{\text{T2T}}$ in this manner, we account for the complexity of generalized bipartite matching and enhance the model's ability to align correspondences across and within modalities.

% %################################################################################
% Figure
\begin{figure}[t!]
\centering
\includegraphics[width=1.0\linewidth]{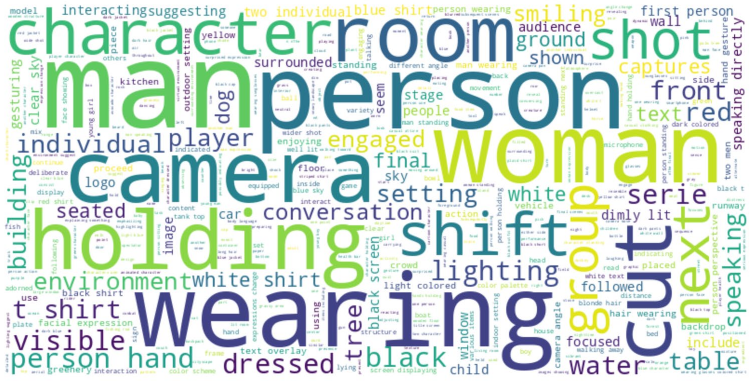}
\vspace*{-0.5cm}
\caption{WordCloud analysis of the SceneWalk dataset.}
\label{fig:5}
\end{figure}
% %################################################################################

\subsection{Word Cloud Analysis.}
The Word Cloud visualization in \cref{fig:5} highlights the richness and diversity of visual cues captured within the SceneWalk dataset. The prominent keywords such as \textit{man}, \textit{woman}, \textit{person}, and \textit{group} reflect the dataset's strong emphasis on human-centric descriptions, focusing on capturing the presence, actions, and interactions of individuals within a scene. These terms highlight that the dataset prioritizes detailed portrayals of people as central subjects, providing rich context about their appearance, activities, and relationships with the surrounding environment.

Furthermore, the inclusion of descriptive spatial and contextual terms (\textit{e.g.,} \textit{stage}, \textit{floor}, \textit{light}, \textit{tree}, \textit{etc},.) illustrates how the dataset prioritizes capturing environmental details alongside subject interactions. This level of granularity ensures that the visual-textual mappings are comprehensive, enabling the dataset to serve as a robust resource for training models that require an in-depth understanding of scene composition and narrative continuity.

By focusing on such fine-grained visual details, the SceneWalk dataset can provide generic scene descriptions, encapsulating nuanced visual content that is critical for multi-modal tasks. The highlighted terms reflect not only the dataset's diversity but also its deliberate emphasis on actionable visual semantics, making it particularly valuable for an intermediate training step, as proposed in \cref{subsec:training}, by enabling models to effectively learn and represent long video knowledge, including scene comprehension and nuanced understanding.

%%%%%%%%%%%%%%%%%%%%%%%%%%%%%%%%%%%%%%%%%%%%%
\begin{table}[b]
    \centering
    \vspace{-0.1cm}
    \setlength\tabcolsep{10pt}
    \resizebox{1.0\linewidth}{!}{\fontsize{18pt}{23pt}\selectfont
        \begin{tabularx}{\textwidth}{p{4.8cm}|*{3}{>{\centering\arraybackslash}X}}
        config & Stage1 & Stage1.5 & Stage2 \\
        \Xhline{1.0pt}
        input modality & image, video & video & video \\
        input frame & \multicolumn{3}{c}{1 FPS} \\
        input resolution & \multicolumn{3}{c}{384 $\times$ 384} \\
        optimizer & \multicolumn{3}{c}{AdamW ($\beta_1, \beta_2{=}0.9, 0.999$)} \\ 
        lr schedule & \multicolumn{3}{c}{cosine decay} \\
        training precision & \multicolumn{3}{c}{BFloat16} \\
        DeepSpeed train & \multicolumn{3}{c}{ZeRO-2} \\
        warmup epochs & \multicolumn{3}{c}{0.03} \\
        trainable params & connectors & full & full \\
        lr\_\{vision, text\} & - & 2e-6 & 2e-6 \\
        lr\_\{LLM, others\} & 1e-3 & 2e-5 & 2e-5 \\
        global batch size & 256 & 8 & 64 \\
        total epochs &  1 & 1 & 1 \\
        Max token drop & 0.0 & 0.7 & 0.4 \\
        \end{tabularx}
    }
    \vspace{-0.3cm}
    \caption{
        Training hyper-parameters for different stages. Here, connectors indicates SR-Router and ST-Connector.
    }
    \label{table:hyperparameters} 
\end{table}
%%%%%%%%%%%%%%%%%%%%%%%%%%%%%%%%%%%%%%%%%%%%%

\section{Training Details of SALOVA}
\paragraph{Training Config.} In this section, we elaborate on the training process of SALOVA-7B. All variations of SALOVA are trained under unified settings, except for the choice of visual encoders. Note, however, that \textit{per-device batch sizes} may vary slightly due to hardware limitations. To equalize the \textit{global batch size} across these variations, \textit{gradient accumulation} is implemented, facilitating a consistent training timeline for each variant. The detailed training configuration for each step can be found in \cref{table:hyperparameters}, which optimizes the use of available GPU memory for batch sizing and ensures efficient training dynamics with limited hardware resource.

\section{Architecture Details of SALOVA}
\paragraph{Network Config.} Here, we explain our network configurations in detail. For the first part of our architecture, the Spatio-Temporal Connector, we employ the Perceiver Resampler~\cite{alayrac2022flamingo} architecture (but, smaller size), which consists of a 2-layer, 2-head Transformer architecture followed by a 2-layer MLP with GELU activation as a projector. For the connector's latent features, we set the number of latent features to 256 and the hidden size to 1024. Next, the second module, the Segment Retrieval Router, consists of a 2-layer, single head Transformer architecture. The Transformer uses a d\_model of 1024 and PReLU as the activation function.

\section{Additional Experiments}

\paragraph{Results of LongVideoBench.} Due to the page limit of the main manuscript, in this additional section, we elaborate on both validation and test set results of LongVideoBench~\cite{wu2024longvideobench} for further demonstration. As in \cref{table:longvbench}, it shows an analogous tendency for Video-MME benchmark, which exhibits a significant performance increase after the short duration video (15s$\sim$). We highlight again that such trend is mainly due to the retrieval capability of SALOVA, which excels in associating visual content with contextual information, even as video lengths increase.

%%%%%%%%%%%%%%%%%%%%%%%%%%%%%%%%%%%%%%%%%%%%%
\begin{table}[h!]
\centering
\resizebox{1.0\linewidth}{!}{
\begin{tabular}{lccccccc}
\Xhline{3\arrayrulewidth} \rule{0pt}{9pt}
                              &                        & \multicolumn{6}{c}{LongVideoBench}                  \\ \cmidrule(lr){3-8}
\multirow{-2}{*}{Model}       & \multirow{-2}{*}{Size} & \rotatebox{90}{8-15s} & \rotatebox{90}{15-60s} & \rotatebox{90}{180-600s} & \rotatebox{90}{900-3600s} & \rotatebox{90}{test set} & \rotatebox{90}{val set}  \\ \hline
\multicolumn{8}{l}{\cellcolor[HTML]{EFEFEF}Proprietary LMMs} \\
GPT-4o~\cite{gpt4o}                        & -                      & 71.6  & 76.8   & 66.7     & 61.6      & 66.7 & 66.7 \\
Gemini 1.5 Pro~\cite{reid2024gemini}                & -                      & 70.2  & 75.3   & 65.0     & 59.1      & 64.4 & 64.0 \\ 
GPT-4-Turbo~\cite{gpt4}                   & -                      & 66.4  & 71.1   & 61.7     & 54.5      & 60.7 & 59.1 \\ \cdashline{1-8}
\multicolumn{8}{l}{\cellcolor[HTML]{EFEFEF}Open-sourced LMMs} \\
VideoChat2~\cite{li2024mvbench}                    & 7B                     & 38.1  & 40.5   & 33.5     & 33.6      & 35.1 & 36.0 \\
VideoLLaVA~\cite{lin2023video}                    & 8B                     & 43.1  & 44.6   & 36.4     & 34.4      & 37.6 & 39.1 \\
PLLaVA~\cite{xu2024pllava}                        & 7B                     & 45.3  & 47.3   & 38.5     & 35.2      & 39.2 & 40.2 \\
LLaVA-1.5~\cite{liu2023improved}                     & 7B                     & 45.0  & 47.4   & 40.1    & 37.0      & 40.4 & 40.3 \\
ShareGPT4Video~\cite{chen2024sharegpt4video}                & 7B                     & \textbf{46.9}  & \underline{50.1}   & 40.0     & 38.7      & 41.8 & 39.7 \\ \cdashline{1-7}
\multicolumn{8}{l}{\cellcolor[HTML]{EFEFEF}Ours} \\
SALOVA-3B & 3B                   & 46.3  & 46.7   & 41.9     & 39.8      & 42.2 & 41.4 \\
SALOVA-3.8B & 3.8B                   & 45.3  & 48.3   & \underline{42.6}     & \underline{40.6}      & \underline{42.9} & \underline{41.6} \\
SALOVA-7$\text{B}^{\dagger}$ & 7B                   & \underline{46.0}  & \textbf{50.7}   & \textbf{44.4}     & \textbf{42.1}      & \textbf{44.5} & \textbf{43.5} \\
\Xhline{3\arrayrulewidth} \rule{0pt}{9pt}
\end{tabular}
}
\vspace{-0.5cm}
\caption{Comparison results for LongVideoBench. The best results are highlighted in \textbf{bold} and the runner-up results are \underline{underlined}. Note that the $\dagger$ mark indicates our efficient size for the frontier model utilizing CLIP~\cite{radford2021learning} as vision encoders (smaller resolution and $144$ visual tokens per frame).}
\label{table:longvbench}
\end{table}
%%%%%%%%%%%%%%%%%%%%%%%%%%%%%%%%%%%%%%%%%%%%%

%%%%%%%%%%%%%%%%%%%%%%%%%%%%%%%%%%%%%%%%%%%%%
\begin{table}[h!]
\centering
\resizebox{1.0\linewidth}{!}{
\begin{tabular}{ccccc}
\Xhline{3\arrayrulewidth}\rule{0pt}{9pt}
\multirow{2}{*}{Ablation} & \multicolumn{4}{c}{Video-MME}                                \\ \cmidrule(lr){2-5}
                          & Short: $\leq$2m    & Mid: 4-15m    & Long: 30-60m   & Overall     \\ \hline
\rowcolor[HTML]{EFEFEF} 
Top-$k$                   & \multicolumn{4}{l}{\cellcolor[HTML]{EFEFEF}: Number of Video Segments for Retrieval}        \\
1                & 48.1          & 44.4	      & 39.1           & 43.9        \\
5                & 48.1          & 45.0	      & 39.2	       & 44.1        \\
9                & 48.3          & 46.3	      & 41.1	       & 45.3        \\ 
13               & 48.1          & 44.7	      & 39.7	       & 44.1        \\ 
\Xhline{3\arrayrulewidth}\rule{0pt}{9pt}
\end{tabular}
}
\vspace{-0.5cm}
\caption{Ablation studies on retrieval number for video segments. We utilize SALOVA-3B model with the CLIP~\cite{radford2021learning} (\texttt{clip-vit-large-patch14-336}) for computational efficiency.}
\label{table:topk}
\end{table}

%%%%%%%%%%%%%%%%%%%%%%%%%%%%%%%%%%%%%%%%%%%%%

\paragraph{Ablation Study for Retrieval Number.} In addition, we conduct an analysis of the number of video segments used for inference on the Video-MME benchmark. In our architectural design, the number of video segments can be dynamically set based on retrieval estimates from the SR-Router, which forwards partial yet pertinent spatio-temporal information from the video to the LMMs. We compare how varying the number of video segments affects performance results. Note that the maximum number of video segments in Video-MME is 13. As in \cref{table:topk}, we clearly observe that increasing the number of frames tends to enhance performance. However, performance saturates after the retrieval number reaches 9. This saturation may be related to the fact that excessive input information becomes more disruptive than helpful for reasoning about partial scenes in the video.

\begin{figure*}[t!]
\centering
\includegraphics[width=1.0\linewidth]{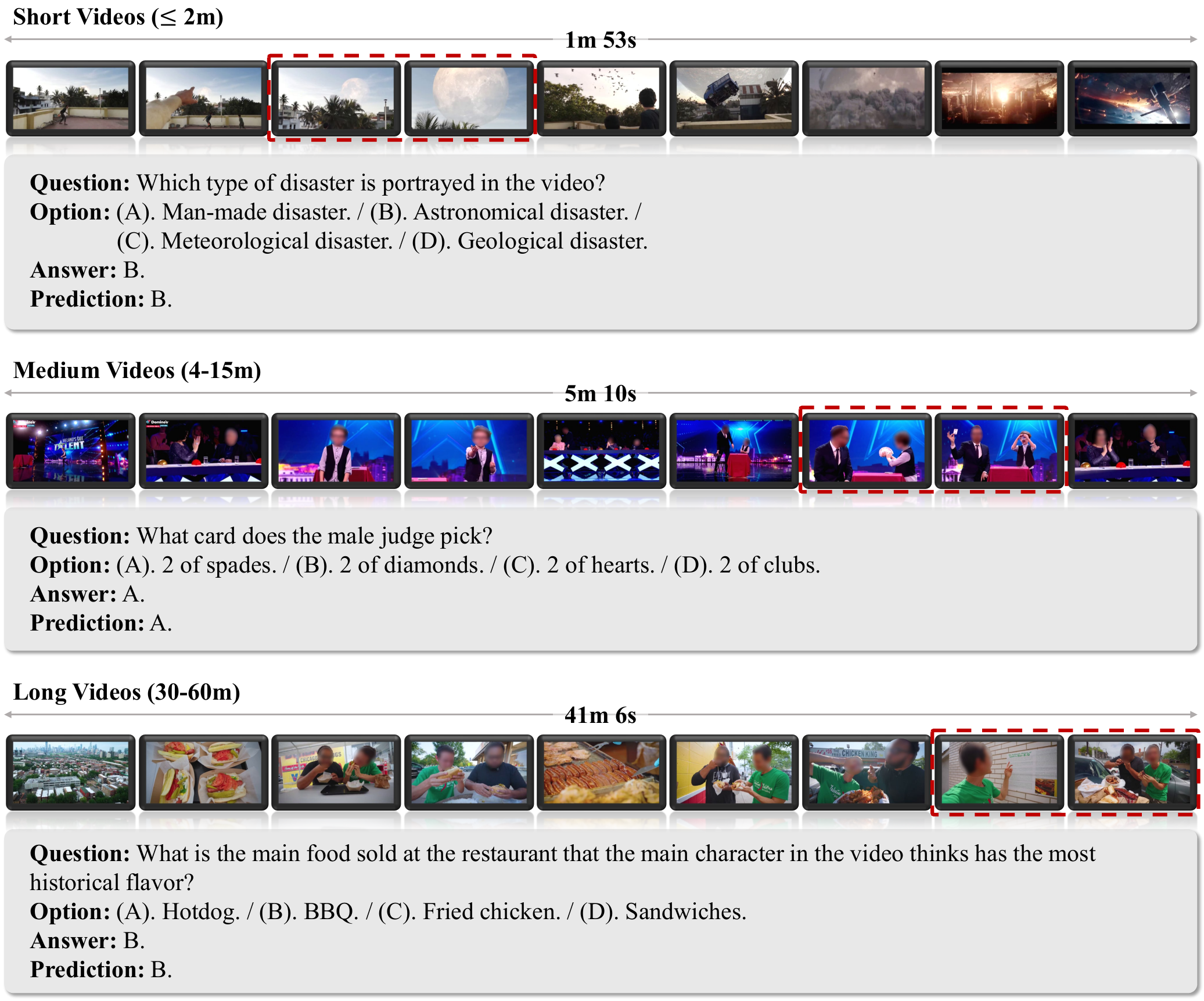}
\vspace*{-0.8cm}
\caption{Qualitative examples on Video-MME~\cite{fu2024video} with SALOVA-7B. Note that the \textcolor[HTML]{C00000}{red dashed lines} indicates the top-1 relevant video segment estimation for the question.}
\vspace*{-0.5cm}
\label{fig:8}
\end{figure*}

\paragraph{Qualitative Results.} We provide qualitative results with varying video lengths to clearly demonstrate the effectiveness of SALOVA across short, medium, and long videos as in \cref{fig:8}. For example, a short video in \cref{fig:8}, our model accurately retrieved a scene of the Moon colliding with the Earth, depicting an astronomical disaster based on the given question. Similarly, in medium-length videos, SALOVA effectively identified the scene where a male judge selects a card corresponding to the question.
Specifically, even in videos longer than 40 minutes, SALOVA accurately identifies scenes related to the correct answer, such as people eating BBQ after exploring the history of the food's origin, based solely on the query input and the video content.

These consistent qualitative results across all lengths indicate that the successful retrieval of pertinent video segments relevant to the input query significantly contributes to the model's efficiency. As demonstrated in our analysis, SALOVA effectively handles different amounts of video data, which supports robust scene understanding and reasoning. 

% %################################################################################
% Figure
% %################################################################################

% %################################################################################
% Figure
% %################################################################################

% %################################################################################
% Figure
% %################################################################################

\end{document}